\documentclass[letterpaper, 10 pt, conference]{ieeeconf}  

\IEEEoverridecommandlockouts                              
\usepackage{tikz}
\usepackage{textcomp}
\usepackage{hyperref}
\usepackage{lipsum}

\newcommand\copyrighttext{%
	\footnotesize \textcopyright 2019 IEEE. Personal use of this material is permitted.
	Permission from IEEE must be obtained for all other uses, in any current or future
	media, including reprinting/republishing this material for advertising or promotional
	purposes, creating new collective works, for resale or redistribution to servers or
	lists, or reuse of any copyrighted component of this work in other works.}
\newcommand\copyrightnotice{%
	\begin{tikzpicture}[remember picture,overlay]
	\node[anchor=south,yshift=10pt] at (current page.south) {\fbox{\parbox{\dimexpr\textwidth-\fboxsep-\fboxrule\relax}{\copyrighttext}}};
	\end{tikzpicture}%
}                                                          
\pdfoutput = 1      
\usepackage{graphicx} 
\usepackage{epstopdf} 
\usepackage{times} 
\usepackage{amsmath} 
\usepackage{amssymb}  
\usepackage{subfiles}
\usepackage{algpseudocode}
\usepackage{algorithm}
\usepackage[caption=false]{subfig}

\newtheorem{thm}{Theorem}

\DeclareMathOperator*{\argmin}{\arg\!\min}
\usepackage{mathtools}
\mathtoolsset{showonlyrefs} 

\title{\LARGE \bf Voluntary Retreat for Decentralized Interference \\ 
	Reduction in Robot Swarms*
}

\author{Siddharth Mayya, Pietro Pierpaoli and Magnus Egerstedt $^{\ddag}$
\thanks{*This work was supported by the Army Research Lab under Grant DCIST CRA W911NF-17-2-0181.}
\thanks{$^{\ddag}$The authors are with the School of Electrical and Computer Engineering at the Georgia Institute of Technology, Atlanta, USA
        {\tt\small \{siddharth. mayya, pietro.pierpaoli, magnus\}@gatech.edu}}
}
\makeatletter
\def\endthebibliography{%
	\def\@noitemerr{\@latex@warning{Empty `thebibliography' environment}}%
	\endlist
}
\makeatother

\begin{document}

\maketitle
\copyrightnotice
\thispagestyle{empty}
\pagestyle{empty}

\begin{abstract}
In densely-packed robot swarms operating in confined regions, spatial interference -- which manifests itself as a competition for physical space -- forces robots to spend more time navigating around each other rather than performing the primary task. This paper develops a decentralized algorithm that enables individual robots to decide whether to stay in the region and contribute to the overall mission, or vacate the region so as to reduce the negative effects that interference has on the overall efficiency of the swarm. We develop this algorithm in the context of a distributed collection task, where a team of robots collect and deposit objects from one set of locations to another in a given region. Robots do not communicate and use only binary information regarding the presence of other robots around them to make the decision to stay or retreat. We illustrate the efficacy of the algorithm with experiments on a team of real robots.
\end{abstract}

\section{INTRODUCTION} \label{sec:intro}
Swarm robotic systems are being increasingly deployed in dynamic real-world environments without the need for a central coordinator overseeing the mission (e.g., see \cite{kitts2008design,csahin2004swarm} and references within). As the size of these swarms increases, robots tend to spend more time and effort avoiding collisions between each other, thereby hurting the overall performance of the swarm \cite{goldberg1997interference}. This phenomenon---an inevitable consequence of the competition for physical space---is called multi-robot interference and has been studied in a wide number of contexts (e.g., \cite{rosenfeld2006study,pini2009interference,hamann2018superlinear}). \par  

In the context of multi-robot \emph{foraging} \cite{castello2016adaptive,shell2006foraging} and \emph{collection}	 tasks \cite{beckers2000fom,martinoli1999probabilistic}, it has been shown that while adding more robots to the swarm at intermediate robot densities increases the total number of objects collected, the increase in performance is sub-linear---caused by a significant decrease in individual robot performance, e.g., \cite{lerman2002mathematical,ostergaard2001emergent}. Owing to the increased operational cost of deploying large robot teams, one would like to obtain the desired overall productivity while utilizing as few robots as possible \cite{hayes2002many}. Achieving such a trade-off is especially challenging in situations where relatively simplistic robots operate in dynamic environments without a central coordinator. \par 

For a swarm of robots executing a \emph{distributed collection} task, this paper develops a decentralized algorithm which enables individual robots to decide between staying in the region and contributing to the task, or retreating from the region when the negative effects of interference outweigh the increased productivity of a larger swarm. In order to allow individual robots to make these decisions, the first part of this paper develops an analytical model which describes the fraction of time that a robot spends performing the primary task as opposed to avoiding collisions with other robots. This allows us to compute an optimal robot density that minimizes a cost function in order to achieve a trade-off between the deployed team size and the overall performance of the swarm. \par

We envision a team of robots collecting objects from randomly scattered ``pick-up" points and depositing them at ``drop-off" points. The locations of these points are unknown to the robots. Such a distributed collection task is of relevance to multi-robot applications such as search \& rescue, mine-clearing, moving waste to recycling stations, transporting passengers to destinations, etc. Furthermore, we assume (i) a constant influx of robots which join the collection task, and (ii) no explicit communication between the robots, thus inspiring the need for a dynamic, yet decentralized mechanism to regulate interference in the domain. \par 

Entomological studies have shown that many social insect colonies possess the ability to regulate interference using decentralized techniques, e.g. \cite{fourcassie2010ant,dussutour2004optimal}. In \cite{aguilar2018collective}, the authors study the ability of \emph{Solenopsis Invicta} ants to regulate densities in narrow tunnels and prevent the formation of flow-stopping clogs. This is partially attributed to the tendency of individual ants to reverse out of crowded tunnels thus reducing the density. Crucially, it has been shown that ants use inter-ant encounters as the primary sensory mechanism to regulate density and perform functions like task allocation and division of labor \cite{gordon2010ant}. \par

Inspired by these observations from biology, \emph{we allow robots to use local inter-robot encounter measurements as the only sensory mechanism to regulate interference}. We extend the ideas presented in our previous work \cite{Mayya-RSS-17}, and demonstrate that not just physical inter-robot collisions but local ``proximity" encounters can allow robots to infer something about the swarm. Using the analytical collision model developed in \cite{mayya2018localization}, we allow robots to measure the density of the swarm by counting interactions. \par

This paper is organized as follows: Section \ref{sec:relatedwork} discusses the contributions of this paper in the context of existing literature. Section \ref{sec:motion_coll} develops analytical models to characterize the frequency of encountering objects and other robots in the domain, given a particular robot density. Section \ref{sec:ctmc} leverages these analytical models to compute the optimal density of robots for achieving a desired trade-off between the size of the swarm and the overall performance. In Section \ref{sec:control}, we develop an algorithm which allows each robot to leave the domain if the measured density of robots in the swarm is above the optimal value. In Section \ref{sec:exp}, the algorithm is deployed on a team of real robots. Section \ref{sec:conc} concludes the paper. 

\section{Related Work}\label{sec:relatedwork}
The modeling and regulation of interference in swarm robotic systems is a widely studied topic (e.g., see \cite{shell2006foraging,hamann2013towards,khaluf2017impact}). The effects of interference on the efficiency of a swarm have been studied experimentally e.g., \cite{rosenfeld2006study,beckers2000fom,arkin1993communication,goldberg2000robust}, and by developing analytical models \cite{lerman2002mathematical,khaluf2017impact}. Techniques to regulate interference range from performing ``aggression" maneuvers to break deadlocks \cite{rosenfeld2006study,vaughan2000go}, partitioning robots among tasks \cite{vaughan2008adaptive}, and team size selection before deploying the swarm \cite{schneider1996study}. \par 
Many biologically inspired techniques allow a group-level division of labor to emerge in order to enhance the efficiency of the swarm, e.g., \cite{castello2016adaptive,bonabeau1998fixed,krieger2000ant}. In the prey-retrieval task studied in \cite{labella2006division}, the authors demonstrate via robotic experiments that allowing robots to modify their tendency to participate in the primary task leads to reduced interference. Similar studies predefine participatory tendencies in robots \cite{krieger2000ant}, or allow them to adjust it online \cite{castello2016adaptive}, demonstrating an improvement in efficiency by running experiments. \par
In contrast to these approaches, this paper develops a decentralized interference reduction mechanism which allows the swarm to adaptively achieve the \emph{optimal} robot density for a desired trade-off between team size and overall productivity. Furthermore, we assume no coordination/communication between the robots and a constant influx of robots joining the task. \par 
Similar to \cite{mather2014synthesis,berman2009optimized}, we use the theory of continuous-time Markov chains (CTMCs) to compute the \emph{distribution} of time spent by robots performing the primary task and in collision avoidance. The development of a CTMC framework is borne out of necessity: in order to allow robots to make decisions at an individual level, we require more than a mean-field characterization of the effects of interference. In the next section, we introduce the problem setup and develop analytical models to describe the interference in the swarm.

\section{Motion and Encounter Models} \label{sec:motion_coll}
\begin{figure}
	\includegraphics[width=0.4\textwidth]{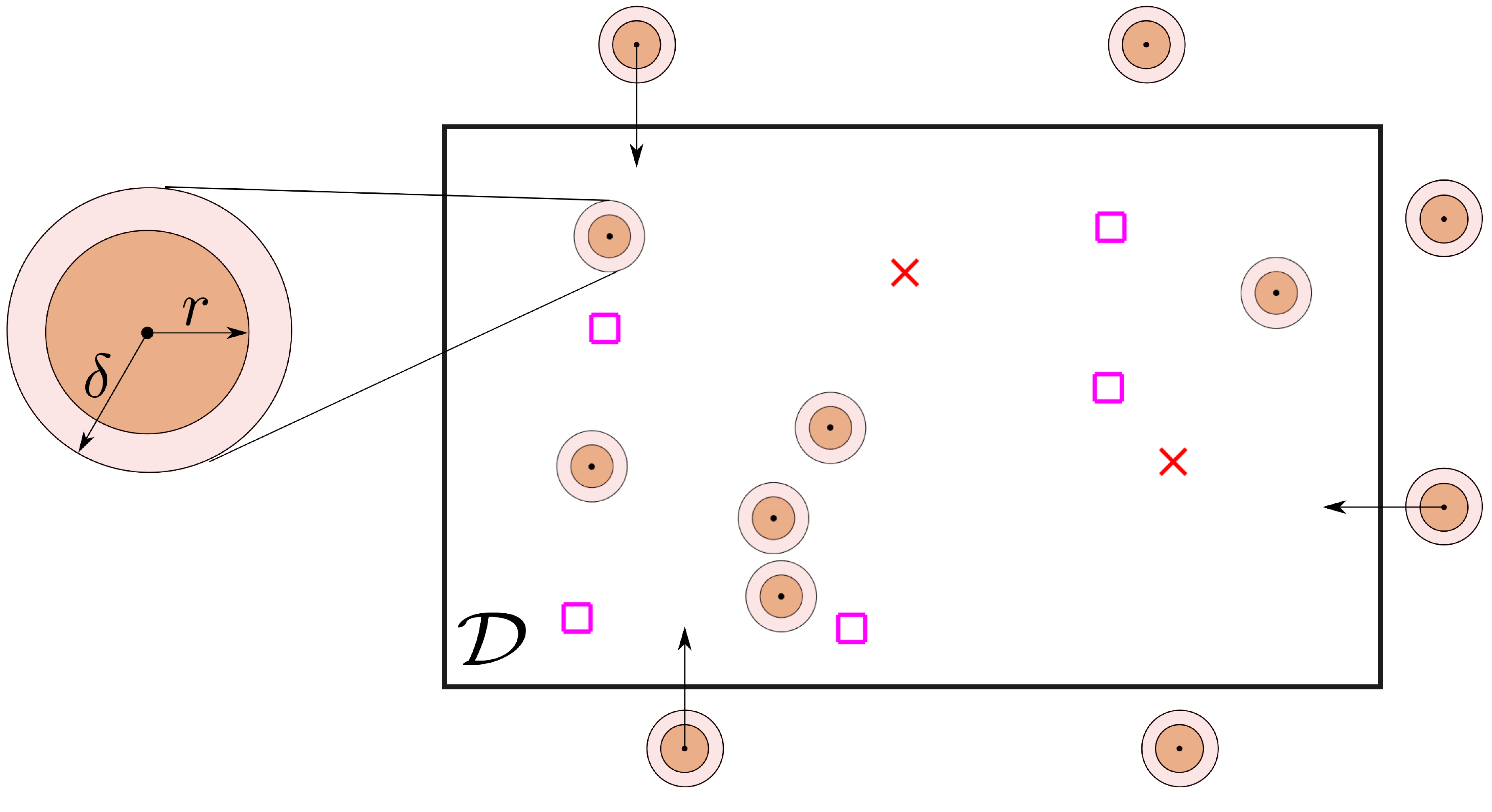}
	\caption{Distributed Collection Task: A team of robots pick up objects from randomly scattered pick-up locations (denoted by $\square$) and deliver them to drop-off locations (denoted by $\times$). An object can be delivered to any drop-off location. Pick-up and drop-off locations are distributed independently with densities $\lambda_p$ and $\lambda_d$, respectively. New robots are stationed outside the domain and enter the domain at a constant rate $\lambda_{in}$, symbolized by the arrows.}
	\label{fig_foraging_rbt}
\end{figure}

We consider a homogeneous team of robots---each with a circular footprint of radius $r$ and a sensing skirt of total diameter $2\delta$ around it to detect objects (see Fig. \ref{fig_foraging_rbt}). Let $v$ denote the average speed of the robots as they move through the domain. \par 

As illustrated in Fig. \ref{fig_foraging_rbt}, the domain $\mathcal{D} \subset \mathbb{R}^2$ contains a set of randomly, but uniformly distributed pick-up locations (marked as $\square$) and drop-off locations (marked as $\times$) from where robots can pick up objects and deliver them, respectively. Each robot picks up or drops off an object if it gets within the sensing distance $\delta$ of the corresponding location. \par 
In this paper, we consider the scenario where picked up objects do not have a specific destination but can be dropped off at any of the drop-off locations. A robot picks up only one object at a time, and delivers it to a destination before picking up another object. We assume that the positions of the pick-up and drop-off points are unknown to the robots, but they know the corresponding non-zero uniform densities, denoted as $\lambda_p$ and $\lambda_d$. Furthermore, new robots initially outside domain $\mathcal{D}$ join the collection task at a constant rate $\lambda_{in}$, as illustrated in Fig. \ref{fig_foraging_rbt}. The rest of this section develops analytical models to characterize the performance of the swarm and the interference at a \emph{particular} robot density, denoted by $\lambda$. \par 

\subsection{Motion Model} \label{sec:motion_model}
The robots do not have any prior knowledge of where the transport locations are, and traverse the domain with trajectories which are ergodic (e.g., \cite{bobadilla2012controlling, shell2005ergodic}) with respect to a uniform distribution. The density distribution $\phi: \mathcal{D} \rightarrow \mathbb{R}$ corresponding to a uniformly ergodic trajectory is given by $\phi(x) = |\mathcal{D}|^{-1}, ~\forall x \in \mathcal{D}$ where $|\mathcal{D}|$ is the area of the domain. This ergodicity property allows us to be general about the motion patterns of the robots---the developed algorithms are applicable as long as the ergodicity conditions are satisfied. Under this ergodicity assumption on the motion of the robots, the distribution of robots will also be uniform in the domain. \par 

\subsection{Inter-Robot Encounters} \label{sec:col_model}
As robots explore the domain, they cross paths with each other. We define two robots $i$ and $j$ as having an ``encounter" when they come within each other's sensing radius, i.e., when $\|p_i - p_j\| \leq 2\delta$ where $p_i$ represents the location of robot $i$. At this point, each robot executes a collision avoidance maneuver (e.g., \cite{fox1997dynamic,wang2017safety}) to resolve the encounter and continues on its intended path. \par 
The encounters experienced by a robot will depend on its motion and the motion of all the other robots in the domain. Similar to our previous work \cite{Mayya-RSS-17,mayya2018localization}, we simplify such complex interactions by making a mean-field approximation and assume that all robots are identical under the effects from other robots in the swarm. There are two relevant questions which arise in this context: how often does a given robot encounter other robots in the domain and how long does it take to resolve an encounter? 

\subsubsection{Inter-Encounter Time Intervals}
\begin{figure}
	\centering
	\includegraphics[width=0.3\textwidth]{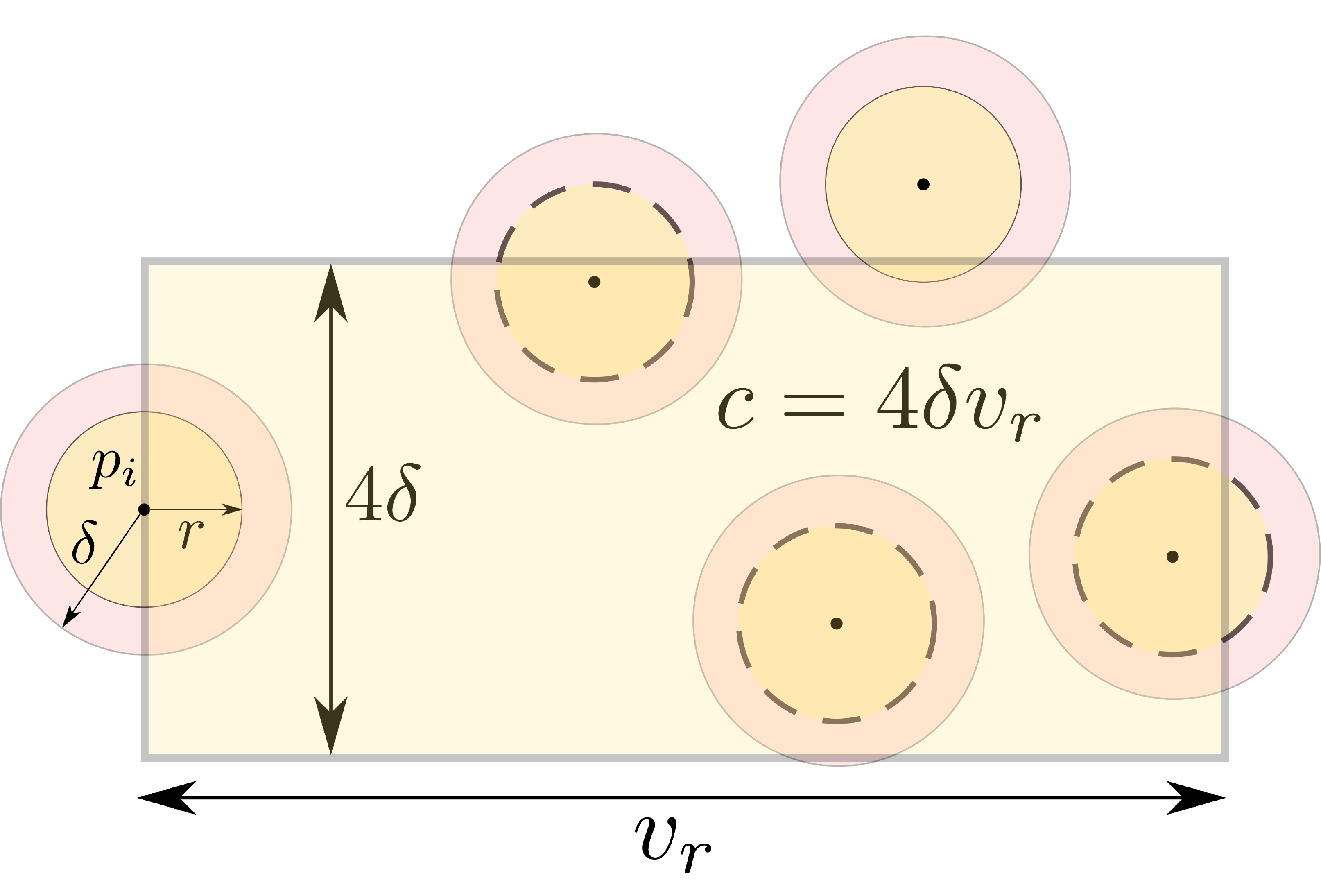}
	\caption{The effective ``sensory" area swept by robot $i$ in unit time is depicted as the shaded region. The expected number of encounters experienced by the robot in unit time will be equal to the number of other robot centers that happen to fall in this region \cite{mayya2018localization,reif2009fundamentals}, whose area is denoted by $c$. Accordingly, the robots experiencing an encounter are indicated with dashed outlines. $v_r$ is the average relative speed between the robots, given by \eqref{eqn_avg_speed}.}
	\label{fig_area_swept}
\end{figure}

Similar to the analysis conducted in kinetic theory of gases \cite{reif2009fundamentals} and in \cite{mayya2018localization,hamann2010space}, if we assume that all robots except the robot under study are stationary, then the expected number of encounters experienced by this robot in unit time will be equal to the number of robot centers which happen to fall into the effective ``sensory" area swept by the robot (see Fig. \ref{fig_area_swept}). This can be computed as $c = 4\delta v$. But, since all other robots are also moving with an average speed $v$, the \emph{effective} speed of the robot can be replaced by the average relative speed between the robots, which can be computed as \cite{mayya2018localization},
\begin{equation} \label{eqn_avg_speed}
v_r = \frac{4}{\pi}v.
\end{equation}
Thus, the effective area swept by the robot is $c = 4\delta v_r$.
Owing to the ergodic motion model developed in Section \ref{sec:motion_model}, the density of robots over the domain $D$ is uniform and is denoted as $\lambda$. Thus, the expected number of encounters experienced by a robot in unit time is given as,
\begin{equation}  \label{eqn_mean_col_tau}
\Omega_c(\lambda) = c \lambda = 4\delta \frac{4}{\pi}v\lambda.
\end{equation}
Although this section characterizes the performance of the swarm for a given robot density $\lambda$, the decentralized control algorithm developed in Section \ref{sec:control} takes into consideration the time-varying densities resulting from the entry and exit of robots. \par 
For a given robot, let $\tau_c(\lambda)$ be the random variable denoting the time interval between entering two successive encounters. Similar to the analysis done in kinetic theory of gases \cite{reif2009fundamentals}, we make the assumption that inter-robot encounters are independent events that are not influenced by each other. This allows us to model $\tau_c(\lambda)$ as an exponentially distributed random variable (see \cite{reif2009fundamentals} for a discussion of this fact),
\begin{equation} \label{eqn_col_tau}
\tau_c(\lambda) \sim exp(\Omega_c(\lambda)).
\end{equation}
In Section \ref{sec:subsec_sim}, we perform simulations to validate the above developed encounter model for robots moving with uniformly ergodic trajectories and executing a specific collision avoidance algorithm.
\subsubsection{Time Spent Resolving an Encounter}
When robots encounter each other, they execute collision avoidance maneuvers. Let $\tau_r$ be the random variable which denotes the time it takes for a robot to resolve an encounter with another robot. $\tau_r$ will depend on a variety of factors such as the collision avoidance maneuvers used by the robots, velocities before collision, presence of domain boundaries, congestion, etc. The uncertainty associated with performing multiple sub-tasks in robotic/vehicular systems has been modeled using sums of exponential random variables \cite{berman2009optimized,russell2008vehicle}. Similar to these formulations, we model $\tau_r$ as an exponentially distributed random variable with a parameter $\rho$ that will vary based on the type of robots and collision avoidance mechanisms used,
\begin{equation} \label{eqn_resolve_col}
\tau_r \sim exp(\rho).
\end{equation}
In practice, the parameter $\rho$ can be estimated by fitting the empirical encounter resolution times to the distribution in \eqref{eqn_resolve_col}. In Section \ref{sec:subsec_sim}, we illustrate using simulations that $\tau_r$ is indeed well-approximated by an exponential distribution for a given collision avoidance algorithm.
\subsection{Object Pick-Up and Drop-Off}
Let $\lambda_p$ and $\lambda_d$ denote the uniform density of object pick-up and drop-off locations in the domain, respectively. As the robots move through the domain, they pick-up/drop-off objects when they enter within a sensing distance $\delta$ of the respective location. Similar to the analysis done in Section \ref{sec:col_model}, the expected number of locations encountered per unit time will be,
\begin{equation} \label{eqn_obj_mean}
\Omega_p(\lambda_p) = 2\delta v \lambda_p ~~;~~ \Omega_d(\lambda_d) = 2\delta v \lambda_d,
\end{equation}
where $\Omega_p$ and $\Omega_d$ correspond to the expectations for pick-up and drop-off locations, respectively. Furthermore, since transport locations are uniformly random, the time elapsed between encountering two such locations can be modeled as an exponentially distributed random variable. Let $\tau_p$ and $\tau_d$ denote the time between encountering two successive pick-up and drop-off locations, respectively,
\begin{equation} \label{eqn_tau_forage}
\tau_p(\lambda_p) \sim exp(\Omega_p(\lambda_p))~~; ~~ \tau_d(\lambda_d) \sim exp(\Omega_d(\lambda_d)).
\end{equation}
The above developed models play a crucial role in the development of a decentralized mechanism to regulate interference, which is the ultimate goal of this paper. We now introduce a simulation setup where the robots solve the distributed collection task while using a particular collision avoidance algorithm.
\subsection{Simulation Setup} \label{sec:subsec_sim}
We validate the theoretical models developed in this paper using a simulated team of differential-drive robots solving the distributed collection task while performing a uniformly ergodic random walk in the domain. The robots deploy minimally-invasive barrier certificates \cite{wang2017safety} to compute safe velocities when they encounter other robots in the domain. \par
\begin{figure}
	\subfloat{\includegraphics[trim = {0.22cm 0cm 9.5cm 0cm}, clip,width=0.24\textwidth]{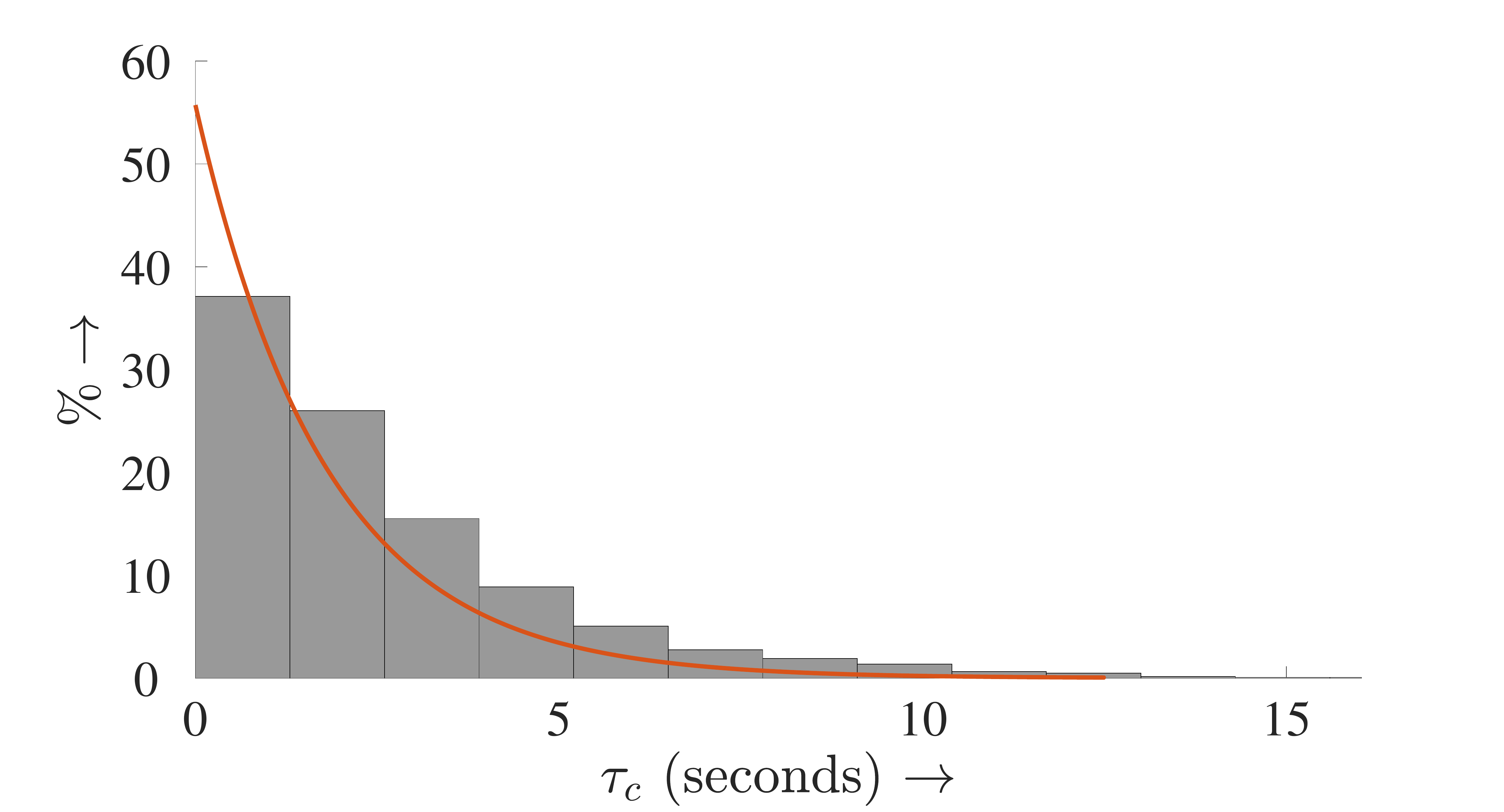}
		\label{subfig_a}
		}
		\subfloat{\includegraphics[trim = {0.22cm 0cm 9.5cm 0cm}, clip,width=0.24\textwidth]{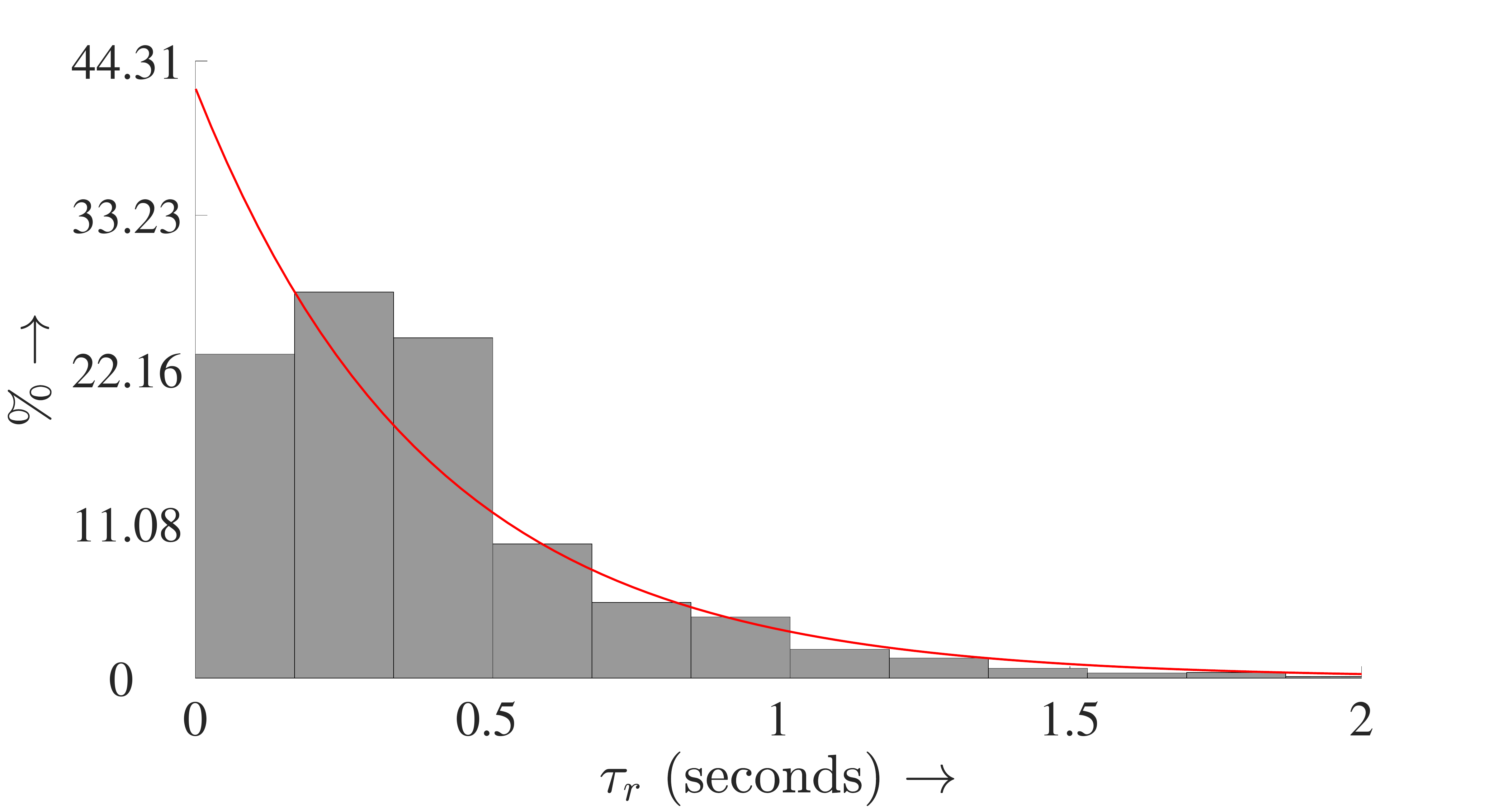}
		\label{subfig_b}
		}
	\caption{Histogram of the inter-encounter time $\tau_c$ and the encounter resolution time $\tau_r$ for a team of simulated robots performing the distributed collection task. In \ref{subfig_a}, the inter-encounter time distribution is well-approximated by an exponential random variable with mean given in \eqref{eqn_mean_col_tau}. \ref{subfig_b} shows the best-fit exponential distribution to the encounter resolution times. Best-fit exponential parameter: $\rho = 2.44$.}
	\label{fig_inter_coll_time}
\end{figure}

For a set of simulation parameters ($\lambda = 0.99$, $\delta = 0.11$, $v = 1$), Fig. \ref{subfig_a} plots the histogram of the empirically obtained inter-encounter times, alongside the theoretical values corresponding to the same density given by \eqref{eqn_col_tau}. Figure \ref{subfig_b} plots a histogram of the time taken by robots to resolve encounters and plots that against a best-fit exponential curve (MATLAB $\texttt{fitdist}$: $\rho = 2.44 $). As seen in both cases, for the given choice of collision-avoidance algorithm, the empirical distributions are well-approximated by the developed encounter models. 

|

\section{Continuous-Time Markov Model}\label{sec:ctmc}
\begin{figure}[t]
	\centering
	\includegraphics[trim = {0cm 0cm 0cm 0cm}, clip,width=0.3\textwidth]{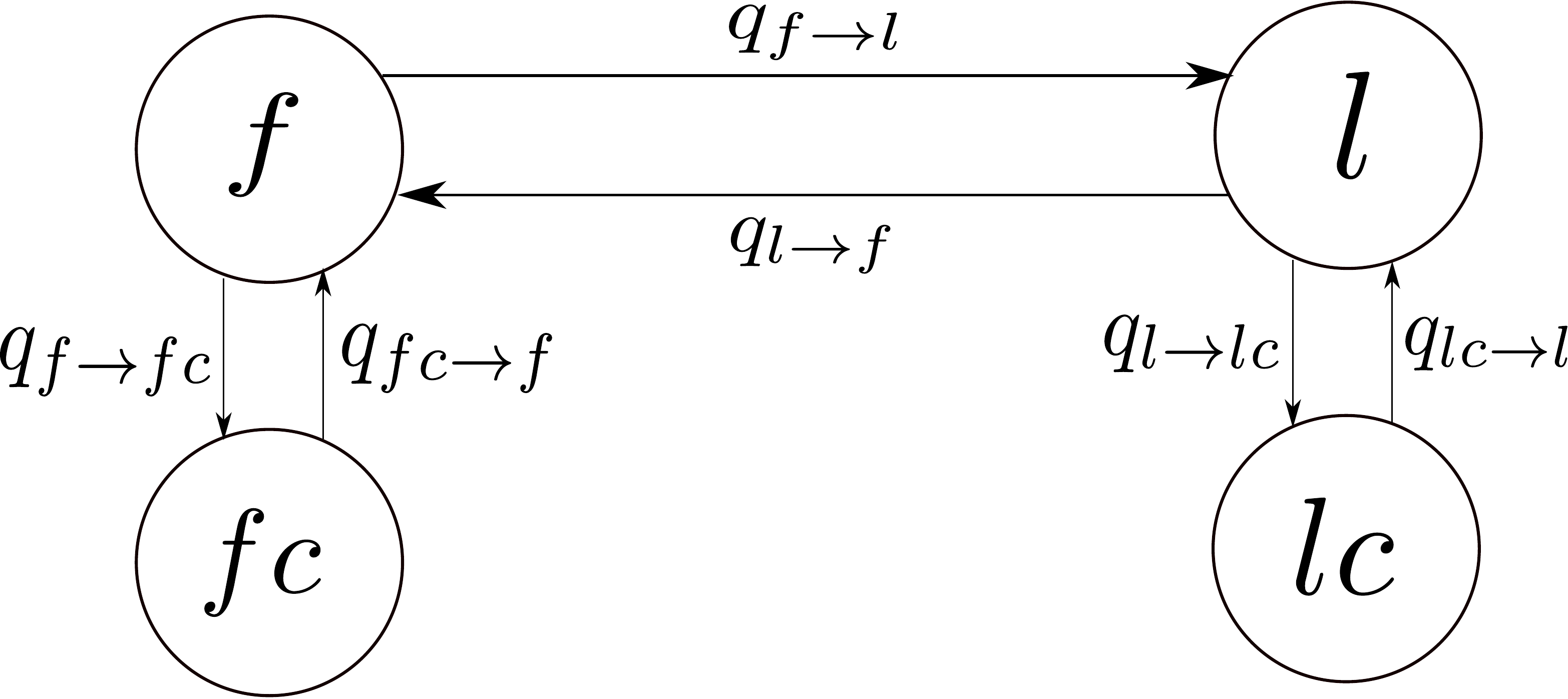}
	\caption{Each robot can be in four different states when performing the \emph{collection} task: free ($f$), free and colliding ($fc$), loaded with an object ($l$), loaded and colliding ($lc$). State transitions are characterized by a continuous-time Markov chain whose transition rates are illustrated.}
	\label{fig_state_dia}
\end{figure}

At a particular time $t$, any robot can be described as belonging to one of four distinct states defined by $\mathcal{V} = \{f,l,fc,lc\}$: free and ready to pick-up an object ($f$), free and colliding with another robot ($fc$), carrying an object ($l$), and colliding while carrying an object ($lc$). The possible transitions between states is illustrated in  Fig. \ref{fig_state_dia}. \par 
Owing to the stochastic nature of the motion of robots and distribution of transport locations, the transition of a robot from one state to another will be stochastic. Let $X(t)$ be the stochastic variable which represents the current state of a chosen robot at time $t$. It does not matter which robot we pick since they are all identical under the mean-field approximation. The memoryless properties of the inter-robot encounter and object-transport models developed in \eqref{eqn_col_tau}, \eqref{eqn_resolve_col}, and \eqref{eqn_tau_forage} imply that the stochastic variable $X(t)$ evolves according to a continuous-time Markov process \cite{karlin2014first} over the possible states $\mathcal{V} = \{f,l,fc,lc\}$.\par 
The probability per unit time of making a transition from state $i \rightarrow j, ~ i,j \in \mathcal{V}$ is characterized by the transition rate $q_{i \rightarrow j}$. Impossible transitions are characterized by $q_{i \rightarrow j} = 0$. The following theorem derives expressions for the transition rates and computes the fraction of time spent by each robot in the different states $\mathcal{V}$.

\begin{thm}
	Consider a team of robots performing the \emph{distributed collection} task as defined in Section \ref{sec:motion_model}. Let $X(t) \in \mathcal{V}$ be the stochastic variable denoting the current state of a given robot at time $t$. Utilizing the collision as well as object transport models developed in \eqref{eqn_col_tau}, \eqref{eqn_resolve_col}, \eqref{eqn_tau_forage}, the fraction of time spent by each robot in a state $j \in \mathcal{V}$ can be computed by evaluating the unique stationary distributions
	\begin{equation} \label{eqn_ss_dist_mc}
	\pi_j = \lim_{t \rightarrow \infty} Prob(X(t) = j), j \in \mathcal{V},
	\end{equation}
	whose expressions are given as,
	\begin{align}
	\pi_f &= \frac{\lambda_d/\lambda_p}{1 + \lambda_d/\lambda_p(1 +\frac{c\lambda}{\rho} (1 + \lambda_p/\lambda_d))}, \\
	\pi_l &= \frac{\lambda_p}{\lambda_d}\pi_f ~;~
	\pi_{fc} = \frac{c\lambda}{\rho}\pi_f ~;~
	\pi_{lc} = \frac{c\lambda}{\rho}\pi_{l} ,
	\end{align}
	where $c = 4\delta v_r$.
\end{thm}
\begin{proof}
The transition rates characterizing the CTMC illustrated in Fig. \ref{fig_state_dia} can be defined as follows.
\begin{itemize}
	\item $q_{f \rightarrow fc}$ and $q_{l \rightarrow lc}$ represent the rates at which a free (or loaded) robot encounters another robot in the domain, which is given by \eqref{eqn_mean_col_tau}.
	\item $q_{fc \rightarrow f}$ and $q_{lc \rightarrow l}$ represent the rates at which a robot resolves an encounter when it is either free or loaded. From \eqref{eqn_resolve_col}, this is simply the parameter $\rho$.
	\item $q_{f \rightarrow l}$ and $q_{l \rightarrow f}$ represent the rate at which a robot picks up objects and drops them off, respectively, given by \eqref{eqn_obj_mean}.
\end{itemize}
Given the transition rates $q_{i \rightarrow j}, i,j \in \mathcal{V}$, we can compute the stationary distribution of the CTMC by constructing the generator matrix $G$ \cite{karlin2014first}, which is given as,
\begin{equation}
G = \begin{bmatrix}
-(c\lambda + m\lambda_p) & m\lambda_p & c\lambda & 0 \\ 
m\lambda_d  &-(c\lambda + m\lambda_d)& 0 & c\lambda \\ 
\rho & 0 & -\rho & 0 \\
0 & \rho & 0  &-\rho
\end{bmatrix},
\end{equation}
where $c = 4\delta v_r$ and $m = 2\delta v$. The stationary distribution $\pi = (\pi_f,\pi_l,\pi_{fc},\pi_{lc})^T$ can be computed by solving the equation $\pi^T G = 0.$
The uniqueness of the stationary distribution $\pi$ follows from the irreducibility and positive-recurrence properties of the embedded discrete-time Markov chain corresponding to the CTMC illustrated in Fig. \ref{fig_state_dia}.
\end{proof} \par
\begin{figure}
	\centering
	\includegraphics[trim = {1.0cm 0cm 4cm 1cm}, clip,width=0.42\textwidth]{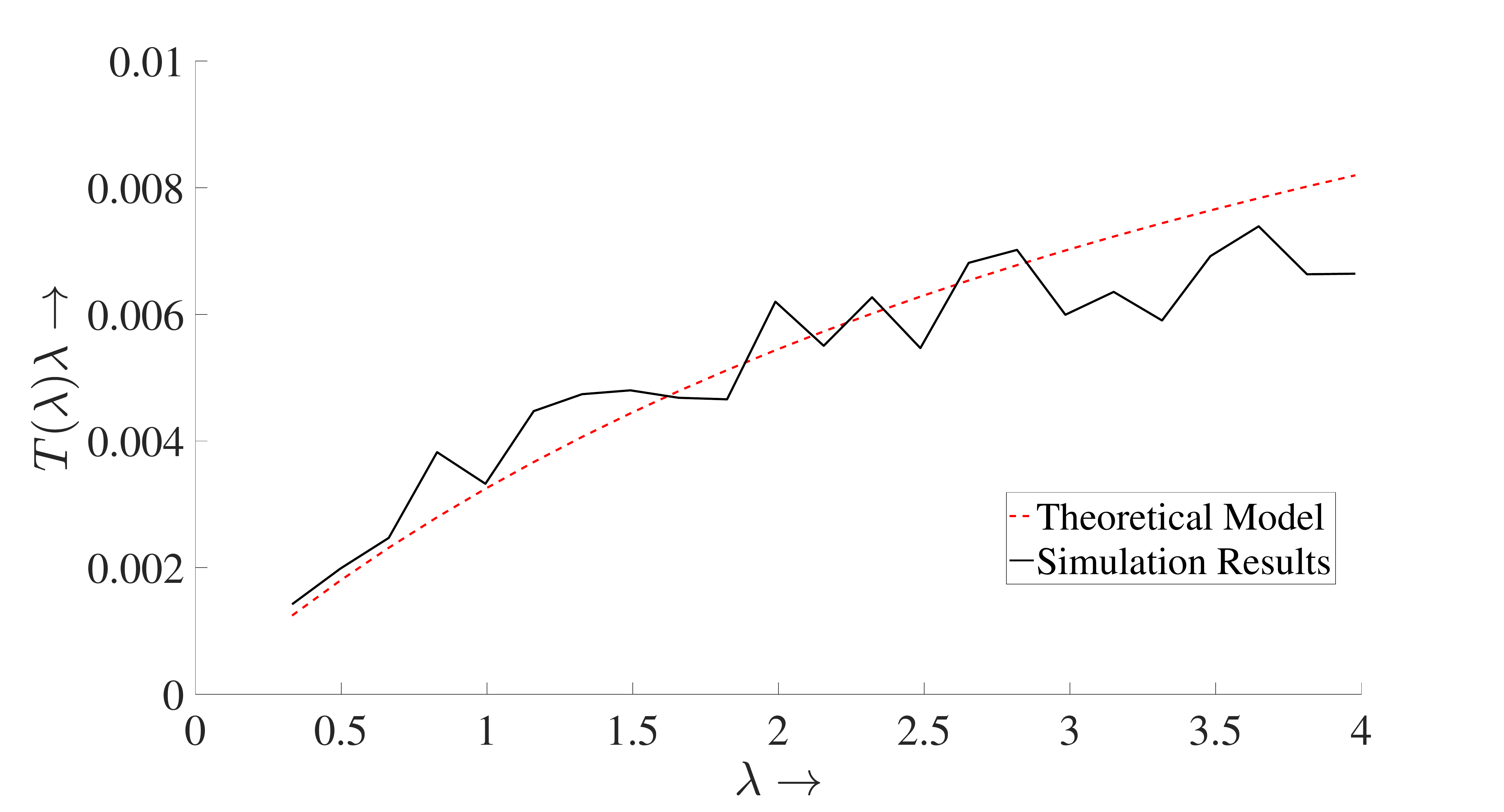}
	\caption{The theoretical evolution of $T(\lambda)\lambda$ as the density of robots (measured in robots$/m^2$) increases is closely matched with the values obtained from simulations with robots employing barrier certificates for collision avoidance. As seen, beyond a certain density, the total output of the swarm increases sub-linearly with the robot density since individual robots spend more time avoiding collisions than transporting objects. This paper is primarily concerned with the performance of the swarm at intermediate densities---the catastrophic jamming occurring at higher densities is not discussed.}
	\label{fig_thruput}
\end{figure}

 
Since $q_{l \rightarrow f}$ represents the rate at which robots deliver objects and become free, the expected number of objects delivered per robot per unit time, denoted by $T$, is given as,
\begin{equation} \label{eqn_thru_rbt}
T(\lambda) = q_{l \rightarrow f} \pi_l = 2\delta v \lambda_d \pi_l(\lambda).
\end{equation}
For the simulation setup introduced in Section \ref{sec:subsec_sim}, Fig.~ \ref{fig_thruput} plots the total number of objects delivered by the robots per unit time per unit area $T(\lambda)\lambda$, against the density $\lambda$ (parameters: $v = 0.1$, $\delta = 0.11$, $\lambda_p = 0.82$, $\lambda_d = 0.66$, $\rho = 2.44$). As seen, the actual productivity of the swarm closely matches the theoretically expected output, barring deviations due to edge effects and the stochastic nature of inter-robot encounters. This paper is primarily concerned with the performance of the swarm at intermediate densities---Fig. \ref{fig_thruput} does not consider very large densities where complete robot jamming can be expected, e.g., \cite{rosenfeld2006study}. \par 
Inspired by the sub-linearity in the total output of the swarm at intermediate densities (see Fig. \ref{fig_thruput}), we would like to compute the optimal density of deployed robots so as to achieve a balance between the swarm size (characterized by $\lambda$) and the total output of the swarm (characterized by $T(\lambda)\lambda$). Such a trade-off can be encoded in the following optimization problem,
\begin{equation} \label{eqn_cost_func}
\argmin_{\lambda \geq 0} J(\lambda) = \frac{C}{T(\lambda) \lambda} + \lambda,
\end{equation}
where $\lambda$ is the density of robots deployed in the domain, and $C$ is chosen to trade-off the cost of deploying robots and the effectiveness of the swarm at performing the task. Setting the gradient of $J(\lambda)$ to zero and solving analytically, we get the optimal density of robots, 
\begin{equation} \label{eqn_optimal_density}
\lambda^* = \argmin_{\lambda \geq 0} J(\lambda) = \sqrt{\frac{C(1 + \frac{\lambda_d}{\lambda_p})}{2\delta v\lambda_d}}.
\end{equation}
In the next section, we allow a robot swarm to autonomously regulate its density to the optimal value by enabling individual robots to retreat from the domain when they sense a higher-than-optimal density.

\section{Decentralized Density Control}\label{sec:control} 
The primary aim of this section is to develop an algorithm that allows individual robots to retreat from the domain when they detect a higher-than-optimal density. The first part of this section develops an ensemble-level control framework for reducing the robot density. A decentralized implementation is later discussed. We also develop an estimation algorithm that allows each robot in the domain to measure the time-varying density of robots $\lambda(t)$. We assume that robots enter and exit the domain by driving through any section of the boundary of domain $\mathcal{D}$ (as depicted in Fig. \ref{fig_foraging_rbt}). 
\subsection{Ensemble Closed-Loop Control}
Given the density of pick-up and drop-off locations $\lambda_p$ and $\lambda_d$, if the density of robots at time $t$, denoted as $\lambda(t)$, is higher than the optimal $\lambda^*$, we would like a certain fraction of the robots to leave the domain (we ignore the degenerate case when $\lambda^* = 0$). Let $\epsilon \lambda$ be the number of robots leaving per unit area per unit time. Then, the population dynamics in the domain can be expressed as,
\begin{equation}\label{eqn_dynamics}
\dot \lambda(t) = \lambda_{in} - \epsilon(\lambda(t)) \lambda(t),
\end{equation}
where $\lambda_{in}$ denotes the known and constant number of robots added to the swarm per unit area per unit time (see Section~\ref{sec:intro}). 
The following choice of $\epsilon$ feedback-linearizes the above system while driving the density of the swarm to $\lambda^*$ using a ``one-sided" proportional controller,
\begin{equation} \label{eqn_eps_control}
\epsilon(\lambda(t)) = 
\begin{cases}
\frac{1}{\lambda(t)}[\lambda_{in} - K_p (\lambda^* - \lambda(t))],\quad \lambda > \lambda^* \\ 
0\qquad \qquad \qquad \qquad \qquad ,~~ \text{otherwise}
\end{cases},
\end{equation}
where $K_p$ is an appropriately chosen proportional gain. A proportional controller was chosen primarily as a simple demonstration of the control mechanism; in general, based on the desired performance requirements, any controller can be used. Applying \eqref{eqn_eps_control} in \eqref{eqn_dynamics}, we get the ensemble closed loop population dynamics,
\begin{equation} \label{eqn_cl_loop_dyn}
\dot \lambda(t) \equiv 
\begin{cases}
K_p(\lambda^* - \lambda(t)), ~~~ \lambda(t) > \lambda^* \\
\lambda_{in}~~~~~~~~~~~~~, ~~ \text{otherwise}
\end{cases}.
\end{equation}
The control parameter $\epsilon(\lambda(t))$ can be interpreted as the probability per unit time with which a robot should leave the domain as a function of the density $\lambda(t)$. Thus, in order for each robot to individually compute $\epsilon$, it must estimate the density of robots in the domain.
\subsection{Estimation}
The robots performing the distributed collection task are not equipped with sophisticated sensors, but can detect encounters with other robots as they move through the domain, as defined in Section \ref{sec:motion_model}. Let $\boldsymbol{e}_i(t)$ denote the total number of robots currently in an encounter with robot $i$. New encounters experienced by robot $i$ can be characterized by an index variable,
\begin{equation}
H_{i}(t) = \begin{cases}
1,  \text{if}~  \exists ~t_1 < t ~ s.t., ~ \forall x \in [t_1,t),~ \boldsymbol{e}_i(x) < \boldsymbol{e}_i(t)  \\
0, ~\text{otherwise}
\end{cases}.
\end{equation}
Denote $y_{iL}(t)$ as the total number of encounters experienced by robot $i$ in a time interval $[t-L,t]$, $t \geq L$,
\begin{equation} \label{eqn_coll_m}
y_{iL}(t) = \int_{t-L}^L H_{i}(t) dt.
\end{equation}
Given this measurement and the encounter model described by \eqref{eqn_col_tau}, the Maximum-likelihood Estimate (MLE) \cite{le1990maximum} of the swarm density for robot $i$ at time $t > L$ is,
\begin{equation} \label{eqn_mle}
\hat \lambda_{i}(t) = \frac{y_{iL}(t)}{4\delta \frac{4}{\pi}vL}.
\end{equation}
The measurement horizon $L$ should be chosen so as to achieve a trade-off between accuracy of estimate and the ability of the robot to track changing densities. 
\subsection{Decentralized Implementation}
Using the maximum likelihood density estimate $\hat\lambda_{i}(t)$, each robot computes its probability of leaving the domain $\epsilon(\hat\lambda_{i}(t))$ using \eqref{eqn_eps_control}. The operations of each robot, executed every $dt$ seconds in software, is illustrated in the following algorithm. \par 
\begin{algorithm}
	\caption{Voluntary Retreat Algorithm}
	\label{alg_sisosig}
	\begin{algorithmic}[1]
		\State{Initialize $\hat \lambda_i = 0, k = 0$}
		\State{Given $\lambda_p$, $\lambda_d$, $\lambda_{in}$, and $C$, compute $\lambda^*$}
		\For{each time $t = k dt$, $t > L$}
		\State{Compute $y_{iL}(t)$ from \eqref{eqn_coll_m} $\leftarrow$ \emph{detecting encounters}} 
		\State{Compute $\hat \lambda_{i}(t)$ from \eqref{eqn_mle}} \label{alg_step_mle}
		\State{Compute $\epsilon(\hat \lambda_i(t)) $ from \eqref{eqn_eps_control}} \label{alg_step_eps}
		\State{Leave the domain with probability $\epsilon(\hat \lambda_i(t)) dt $}
		\State{$k = k + 1$}
		\EndFor
	\end{algorithmic}
\end{algorithm}
Before deployment, each robot computes the optimal density $\lambda^*$ using the density of transport locations and influx rate of robots. After an initial wait of $L$ seconds (during which time robots are collecting encounter information), each robot computes the MLE estimate in step \ref{alg_step_mle} and the resulting $\epsilon$  in step \ref{alg_step_eps}. Finally, the robot flips a biased coin to leave the domain with a probability $\epsilon(\hat \lambda_i(t)) dt$. \par 
Each robot $i$ will have a different estimate of the density; consequently, $\epsilon(\hat\lambda_{i}(t))$ will vary from one robot to another. While this implies that robots behave differently from each other, we demonstrate in the next section that, for a team of real world robots, the algorithm indeed allows the robots to regulate the density of the swarm to the optimal value.

\section{Experimental Results} \label{sec:exp}
\begin{figure}[t]
	\centering
	\vspace{0.7em}
	\subfloat[]{\includegraphics[trim = {0cm 0cm 0cm 0cm}, clip, width=0.40\textwidth]{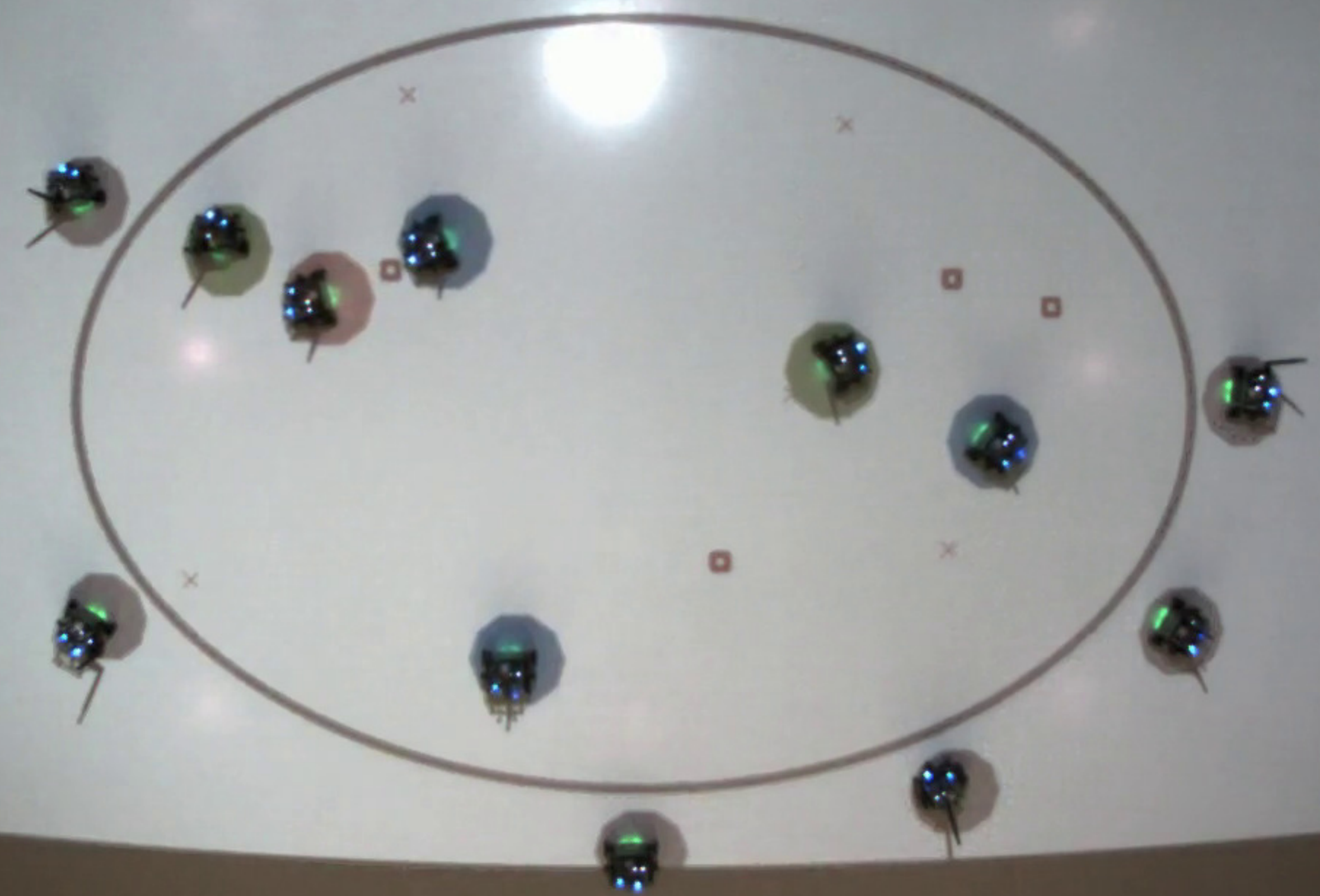}
	\label{fig_exp}
	} 
	\\
	\subfloat[]{\includegraphics[trim = {2cm 0cm 4.2cm 1cm}, clip, width=0.43\textwidth]{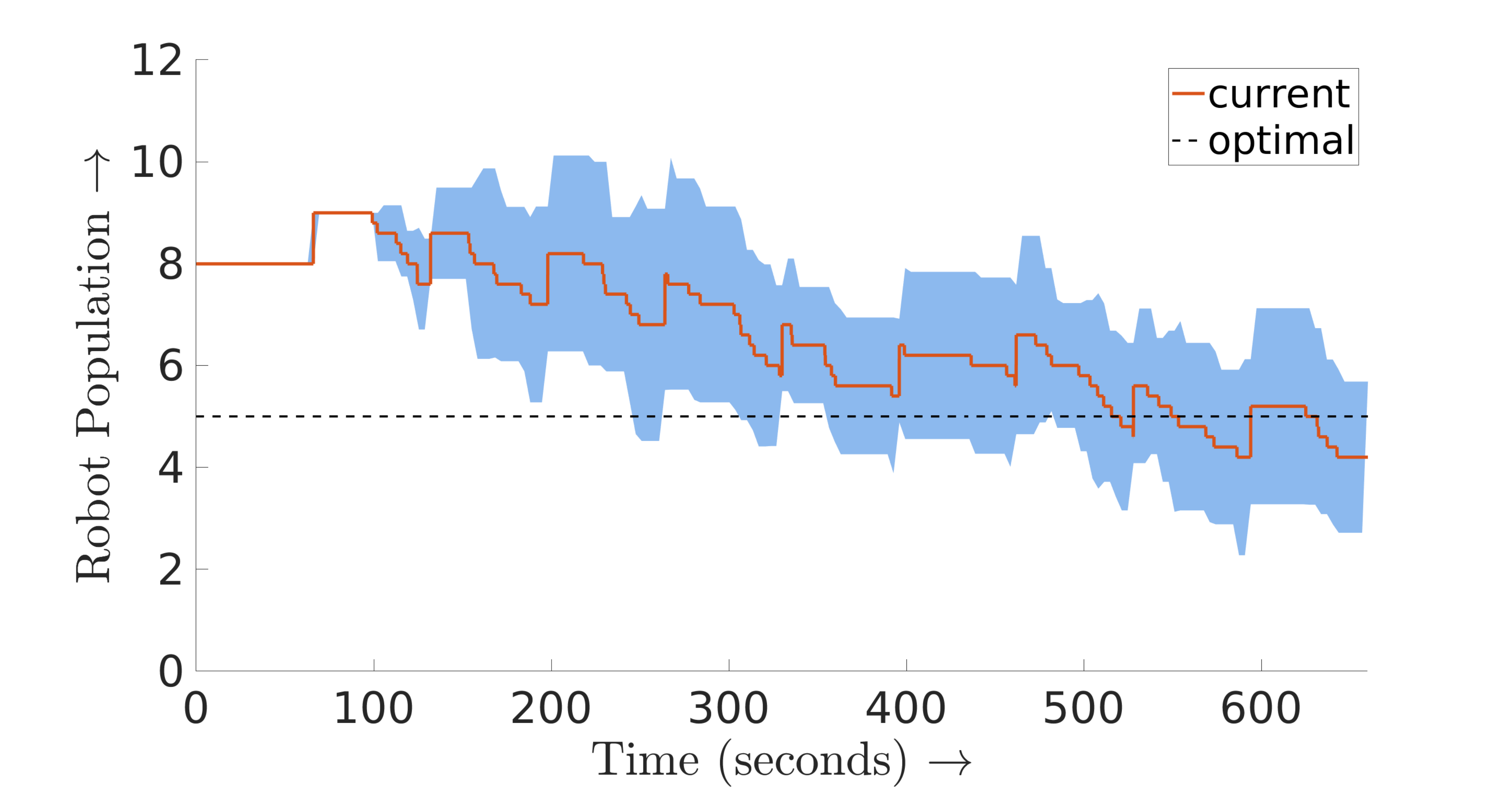}
		\label{fig_graphs}
	}
	\caption{Experimental verification of the Voluntary Retreat Algorithm on a team of 12 real robots operating on the Robotarium. An overhead projector is used to project additional information onto the robot arena. The experiment begins with 8 robots inside the elliptical domain seen in \ref{fig_exp}. Robots stationed outside enter at a steady rate $\lambda_{in}$. Blue and red circles projected around each robot signify \emph{free (f)} and \emph{loaded (l)} robots, respectively. Green circles indicate that a robot has decided to retreat from the domain. These robots drive out of the elliptical domain via any section of the boundary. The robot population data in \ref{fig_graphs} is averaged over 5 experimental runs to demonstrate the consistent performance of the algorithm, with one standard deviation depicted by the shaded region.}
\end{figure}

The developed algorithm was deployed on the Robotarium \cite{pickem2017robotarium}, a remotely accessible swarm robotics testbed. The experiment was initialized with 8 robots starting inside the elliptical domain (with $1.2m$ and $0.8m$ as the semi-major and semi-minor axis length) shown in Fig. \ref{fig_exp}, deploying minimally-invasive barrier certificates \cite{wang2017safety} for collision avoidance. An additional $4$ robots were initially placed outside and entered the domain at a steady rate. Pick-up ($\square$) and drop-off ($\times$) locations are marked on the domain. \par
For the following parameters: $\lambda_p = 1.32$, $\lambda_d = 1.32$, $C = 0.03$, $\delta = 0.1$, $v = 0.1$, each robot computes the optimal swarm density using \eqref{eqn_optimal_density}. The robots perform an unbiased random walk in the domain while encountering other robots. Using Algorithm \ref{alg_sisosig} (with parameters $L = 90$, $dt = 0.03$, $K_p = 0.03$, $\lambda_{in} = 0.02$), each robot flips a biased coin to decide whether it should stay or retreat.\par 
The experiment was repeated 5 times to average the results, and demonstrate the consistent performance of the voluntary retreat algorithm. Figure \ref{fig_graphs} plots the mean of the true robot population against the optimal value, along with the standard deviation (represented by the shaded region). As seen, the rate at which robots retreat reduces as the true population approaches the optimal value, in accordance with the closed-loop dynamics given in \eqref{eqn_cl_loop_dyn}. Even with variations in the density estimates of the robots as well as the constant influx of new robots (indicated with upwards jumps in Fig. \ref{fig_graphs}), the closed loop algorithm allows the population to settle around the optimal value. 
\section{Conclusion} \label{sec:conc}
The technique presented in this paper allows a robot swarm to autonomously curb the negative effects of spatial interference by enabling individual robots to either stay and perform the distributed collection task, or retreat so as to reduce the density. Each robot uses only binary information regarding the presence of other robots around it to estimate the density of the swarm, and decides whether to stay or voluntarily retreat from the domain. We demonstrate that the swarm regulates the density to an optimal value, which is analytically computed to achieve a trade-off between swarm size and overall productivity. Multi-robot experiments demonstrate the efficacy of the developed algorithm.
\vspace{2em}
\addtolength{\textheight}{-2.4cm}
\bibliographystyle{IEEEtran}

\bibliography{references}

\end{document}